\documentclass[journal]{IEEEtranTIE}
\usepackage{graphicx}
\usepackage{cite}
\usepackage{picinpar}
\usepackage{amsmath}
\usepackage{url}
\usepackage{flushend}
\usepackage[latin1]{inputenc}
\usepackage{colortbl}
\usepackage{soul}
\usepackage{multirow}
\usepackage{pifont}
\usepackage{color}
\usepackage{alltt}
\usepackage[hidelinks]{hyperref}
\usepackage{enumerate}
\usepackage{siunitx}
\usepackage{breakurl}
\usepackage{epstopdf}
\usepackage{pbox}

\usepackage[caption=false,font=footnotesize]{subfig}

\usepackage{xcolor}
\usepackage{soul}
\sethlcolor{cyan}

\usepackage[flushleft]{threeparttable}

\begin{document}
\title{Explainable Machine Learning for Multilayer Planar Winding Inductance Estimation}

\author{
	\vskip 1em
	
	Spyros Rigas, %\emph{Student Membership},
	Theofilos Papadopoulos, %\emph{Membership},
	Georgios Alexandridis,
	and Antonios Antonopoulos %\emph{Membership}

	\thanks{
	
		%Manuscript received Month xx, 2xxx; revised Month xx, xxxx; accepted Month x, xxxx.
		%This work was supported in part by the xxx Department of xxx under Grant  (sponsor and financial support acknowledgment goes here).
		
		Theofilos Papadopoulos and Antonios Antonopoulos are with the School of Electrical and Computer Engineering, National Technical University of Athens, Greece, (e-mail: teopap@mail.ntua.gr). 
		
		Spyros Rigas and Georgios Alexandridis are with the Department of Digital Industry Technologies, National and Kapodistrian University of Athens, Greece (corresponding author e-mail: spyrigas@uoa.gr).
	}
}

\maketitle
	
\begin{abstract}
Rapid and accurate self-inductance estimation for multilayer rectangle-shaped planar windings is essential for modern high-frequency power converters, yet traditional workflows rely on complex mathematical equations, rigid monomial formulas or unexplainable black-box machine learning (ML) models that degrade severely outside their training domain. This paper introduces an explainable ML framework unifying post-hoc feature attribution (SHAP and permutation importance) with Kolmogorov--Arnold Network-guided symbolic regression via the SR-KAN framework to discover closed-form analytical equations without prior structural assumptions. Evaluated on a new open-source dataset of over 10,000 Finite Element Analysis (FEA) simulations across seven out-of-distribution (OOD) classes, standard tree-based ensembles exhibit severe extrapolation errors ($\boldsymbol{>}$ 36\%), whereas the unconstrained SR-KAN expression achieves a robust OOD relative error of 8.22\%. Experimental verification across 55 physical printed circuit board prototypes (up to 8 layers, with inductances from 4.11~$\boldsymbol{\mu}$H to 559.27~$\boldsymbol{\mu}$H) confirms that the KAN-discovered expression translates effectively to real-world hardware, predicting inductance with a mean absolute relative error of 6.26\%. To support reproducible research, the complete FEA simulation dataset and prototype measurements are released open-source.
\end{abstract}


%\markboth{IEEE TRANSACTIONS ON INDUSTRIAL ELECTRONICS}%
{}

\definecolor{limegreen}{rgb}{0.2, 0.8, 0.2}
\definecolor{forestgreen}{rgb}{0.13, 0.55, 0.13}
\definecolor{greenhtml}{rgb}{0.0, 0.5, 0.0}

\section{Introduction} \label{sec1}

% % Theofilos + Spyros

\IEEEPARstart{W}{ith} the continuous push toward high-power-density, miniaturization, and high-efficiency for modern power electronics, planar magnetics have emerged as a viable alternative to bulky conventional wire-wound components. With printed circuit board (PCB) copper traces, planar windings (PWs) offer highly-repetitive parameters, easy and inexpensive structural reproducibility, and low profiles ideal for high-frequency operations \cite{overview_wouters25}. These attributes have made PWs attractive in power applications, including electric vehicle (EV) powertrains \cite{ev_lee24}, wide-bandgap (WBG) semiconductor high-frequency converters \cite{wbg_barz25}, data-center power converters \cite{integration_nahib24}, and even gate-driver power supplies \cite{gd_yan25}. 

Particularly in resonant topologies, such as LLC, CLLC and Dual Active Bridge (DAB) converters, the precise estimation of magnetic parameters is important, as variations directly affect soft-switching conditions and shift the resonant frequency \cite{llc_zhang20}. To address this, magnetic integration methodologies are frequently deployed to take advantage of leakage and magnetizing inductances within a singular structural layout, shrinking the overall converter footprint \cite{integration_liu23, integration_liu26}. While many integrated structures utilize a ferrite core, giving special consideration to windows or air-gaps, to control flux distributions \cite{core_liu23}, coreless (air-core) planar implementations can be also used, where linear operation, zero core losses, reduced weight, and cost minimization are prioritized \cite{llc_ho20}.

Regardless of the core configuration, achieving a highly accurate and computationally inexpensive estimation of winding inductance remains an open issue. Traditional design workflows heavily rely on numerical Finite Element Analysis (FEA), which yields high accuracy but demands prohibitive computational time and meshing overhead when executing large-scale geometric optimization loops. Conversely, classical analytical expressions (e.g., Wheeler, Rosa, or standard Monomial equations) offer instant evaluation but exhibit unacceptable accuracy degradations as the winding deviates from the standard square-shape, single layer design. Others have proposed estimation equations based on geometrical parameters, but for radio-frequency (RF) small-dimension designs \cite{aebischer20}.

To bridge the gap between slow and expensive numerical simulations and rigid analytical formulas, FEA and data-driven results have been introduced to the domain of planar magnetics. For instance, Machine Learning (ML) models have been applied to optimize the highly non-linear parameter spaces of medium-frequency transformers in solid-state topologies \cite{ml_sst_noh26}, while advanced neural networks have been utilized to map the mutual inductance variations in dynamic wireless power transfer setups \cite{nn_wpt_boulanger25}. While such approaches can fit complex datasets with near-perfect accuracy, they operate as black boxes that obscure the underlying physical mechanisms.

In engineering, explainability is a practical necessity for understanding the core mechanics of the system, its sensitivity to the parameters, and ensuring safety margins and stability boundaries. For these reasons, our previous work \cite{estimation_pap25} proposed a Multiple Linear Regression (MLR) framework to derive a closed-form monomial equation for specific multilayer arrangements. Although this MLR formulation yielded an inherently explainable expression, purely data-driven non-explainable models can generally achieve superior estimation accuracy. Furthermore, deriving that analytical formula relied heavily on prior domain knowledge -- specifically, assuming that a power-law product would describe the problem well -- which limits its generalizability when no prior hint of an explicit symbolic form exists.

To achieve explainability without sacrificing predictive accuracy, this paper proposes a holistic explainable ML framework for planar magnetics that unifies two methodological paradigms: post-hoc attribution and Symbolic Regression (SR). The first relies on post-hoc explainability, where attribution techniques such as SHapley Additive exPlanations (SHAP) \cite{shap} and permutation importance \cite{permutation} are applied to pre-trained black-box models, including tree-based ensembles \cite{RFs, xgboost} and support vector regressors (SVRs) \cite{SVR}. The second paradigm bypasses the black box by searching directly for analytical, closed-form expressions without making prior structural assumptions about the modeled system. To this end, we utilize SR-KAN \cite{SRKAN}, which leverages the interpretable univariate edge functions of Kolmogorov--Arnold Networks (KANs) \cite{KAN}, and systematically benchmark it against the accepted state-of-the-art in evolutionary symbolic regression, PySR \cite{pysr}, as well as the original MLR methodology of \cite{estimation_pap25}.

The proposed framework is deployed on a newly developed, open-source dataset comprising over 10,000 FEA simulations of Multilayer Rectangle-Shaped Planar Windings (MLRPWs), which we release as a core contribution of this work \cite{zenodo_db}. This dataset captures geometric variations across extensive layer counts and consists of a standard training and testing subset alongside separate, independent edge cases explicitly categorized into distinct classes of out-of-distribution (OOD) configurations. Moreover, to validate the accuracy of the simulated dataset under real-world laboratory tolerances, we manufacture and measure more than 50 physically printed MLRPW prototypes across diverse layer configurations. Crucially, while winding inductance estimation serves as the workhorse application in this study, the introduced explainable pipeline is entirely domain-agnostic and can be readily generalized to alternative modeling tasks across industrial electronics.

The remainder of this paper is organized as follows: Section \ref{sec2} introduces the geometric parameters of MLRPWs and outlines the open-source FEA dataset generation, including specific geometric configurations selected for OOD testing. Section \ref{sec3} details the explainable ML methodology, highlighting both core aspects of the proposed framework: post-hoc explainability via SHAP and permutation importance, as well as equation discovery via SR-KAN and other symbolic regression baselines. Section \ref{sec4} presents the benchmark evaluations and post-hoc attributions across diverse ML models and the predictive performance of the derived analytical expressions across both standard and OOD domains. Section \ref{sec5} provides experimental laboratory verification against more than 50 physically fabricated MLRPW prototypes, and Section \ref{sec6} concludes the paper.

\section{Geometric Parameters and Dataset explanation} \label{sec2}

% % Theofilos

\begin{figure}[!b]\centering
	\centering
	\subfloat[]{%
		\includegraphics[width=0.85\columnwidth]{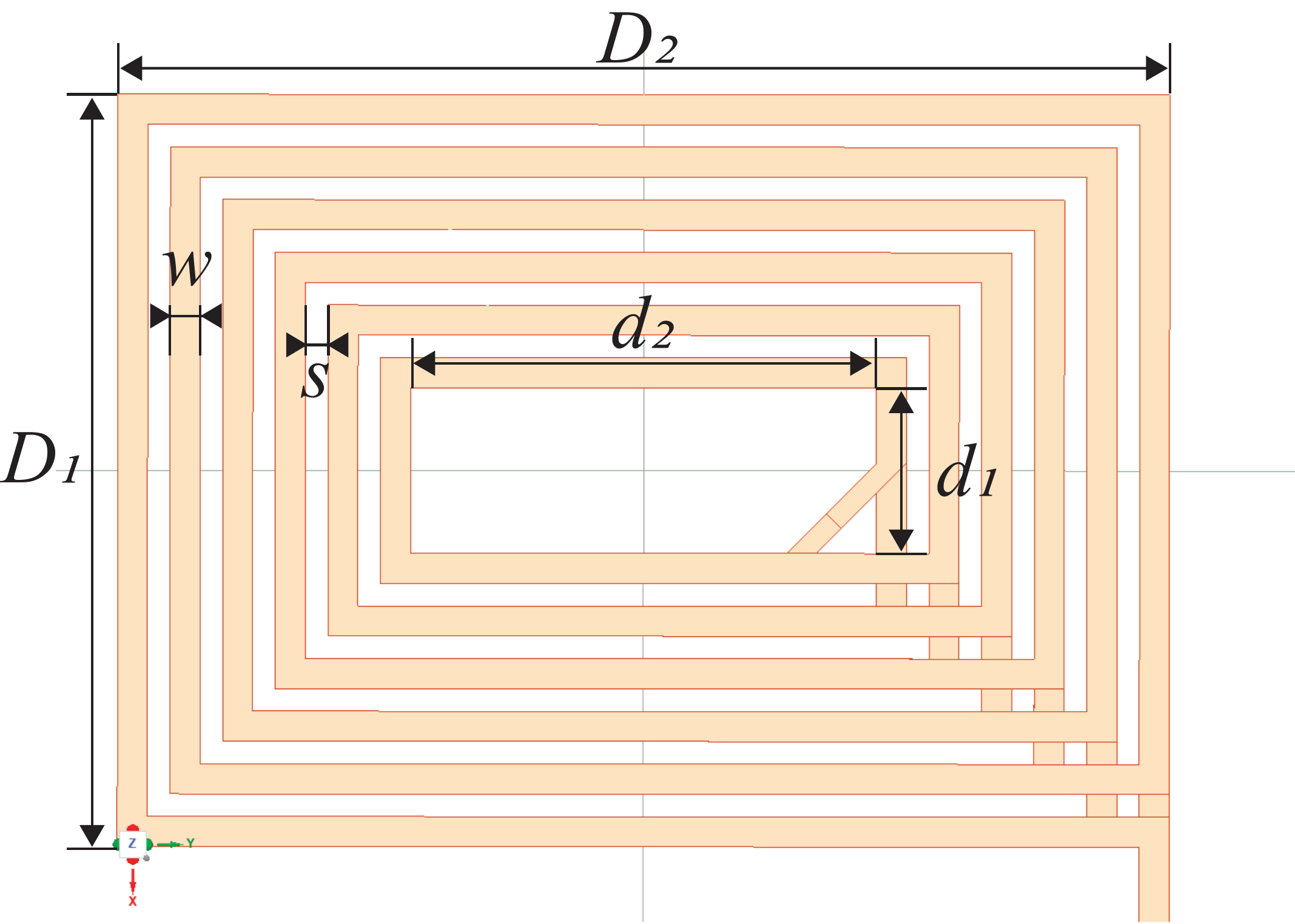}
		\label{fig:case_a}
	}
	\vspace{2mm}
	\subfloat[]{%
		\includegraphics[width=0.75\columnwidth]{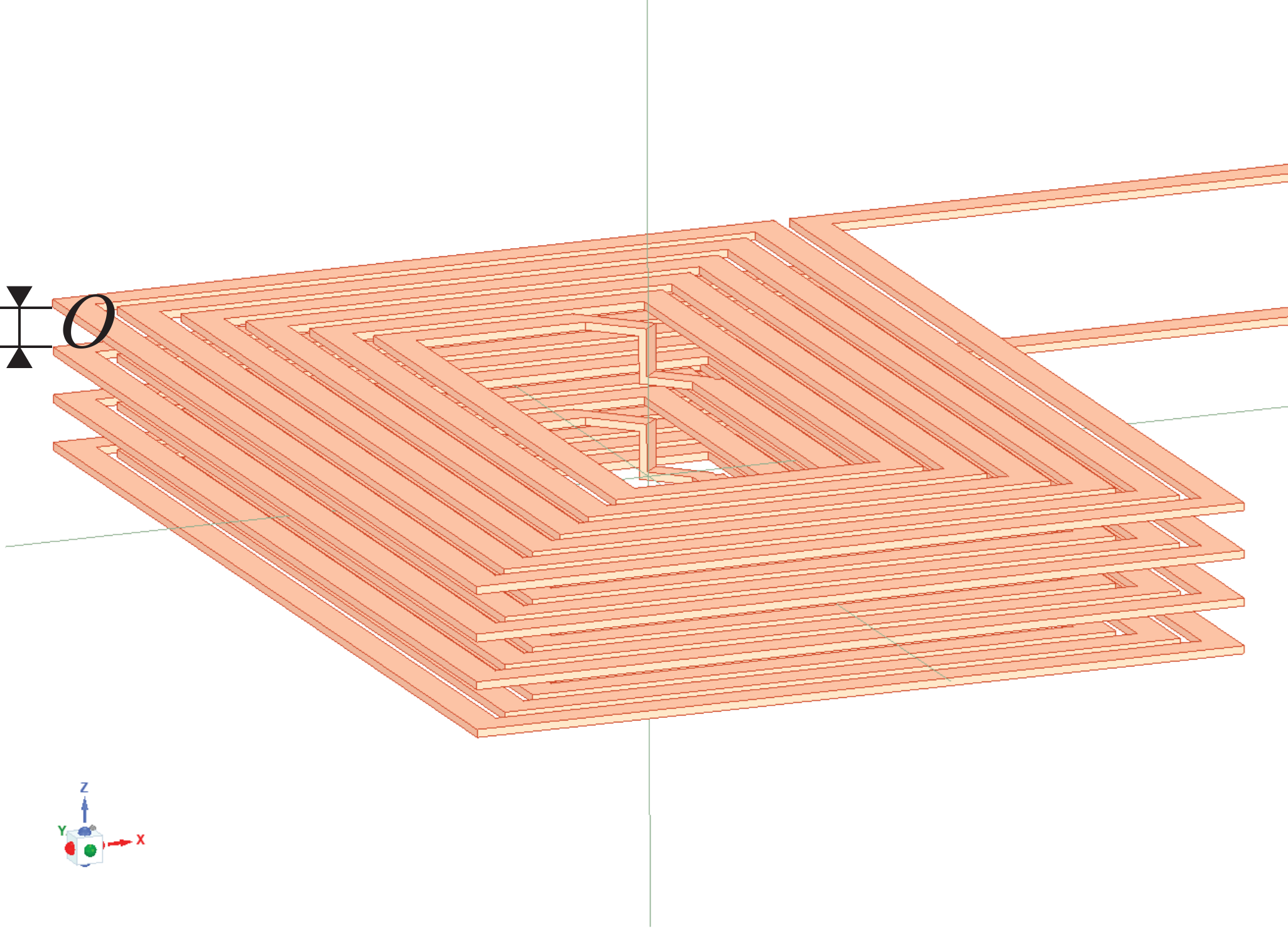}
		\label{fig:case_b}
	}
	\caption{Geometric parameters for a typical MLRPW (4-layered).}
	\label{fig:two_cases}
\end{figure}

The self-inductance of an air-core winding is dictated by its physical shape and geometric proportions. For MLRPWs, this behavior is defined by a multi-dimensional parameter space containing both coil-level and trace-level variables. As shown in Fig. \ref{fig:two_cases}, the complete physical layout is determined by the outer-side lengths $D_1$ and $D_2$, the number of turns per layer $N_T$, the total number of layers $N_L$, the copper trace width $w$, the horizontal inter-turn trace spacing $s$, and the vertical distance between consecutive insulating layers $O$. From these independent variables, the inner-aperture side lengths $d_1$ and $d_2$ are explicitly calculated as:

\begin{equation}
	\label{eq:d}
	d_i = D_i - 2 N_T (w+s) + 2s
\end{equation}

\noindent where $i \in \{1, 2\}$ denotes the respective orthogonal axes of the rectangular profile.

Planar windings can span vastly different physical scales depending on their targeted industrial application. While small-scale profiles on the micrometer scale are common in low-power RF chips, power applications require much larger footprints to satisfy the thermal and the electrical conditions. In this direction, the geometric parameters in this study are chosen based on the IPC-2221 design standard. The trace width $w$ spans from 3~mm to 5~mm, which corresponds to a continuous current-carrying capability of approximately 5~A to 10~A for a 20~$^\circ$C maximum temperature rise limit. Similarly, the horizontal trace spacing $s$ ranges from 0.1~mm to 0.5~mm, satisfying the 300~V turn-to-turn insulation voltage-withstand requirements for permanent polymer-coated external conductors under IPC class B4. To accommodate real-world magnetic cores (such as standard EE or EI geometries), the inner dimensions $d_1$ and $d_2$ are restricted to a minimum clearance threshold of 17~mm. The complete boundaries of the core design space are summarized in Table \ref{tab:dims}.

\begin{table}[!b]
	\renewcommand{\arraystretch}{1.3}
	\caption{Value Ranges for Dataset Parameters}
	\centering
	\label{tab:dims}
	%\centering
	%	\resizebox{\columnwidth}{!}{
		\begin{threeparttable}
		\begin{tabular}{c c l c c}
			\hline\hline \\[-3mm]
			\textbf{Parameter} & \textbf{Value Range} &  & \textbf{\# of Values} & \textbf{Comments} \\[0.5ex] \hline \\[-3mm]
			$D_1$ & \multicolumn{1}{c}{\multirow{2}{*}{70:10:160\tnote{1}}} & mm & 10 & \multicolumn{1}{c}{\multirow{2}{*}{$D_1 \leq D_2$}} \\
			$D_2$ &  & mm & 10 & \multicolumn{1}{c}{} \\
			$d_1$ & \multicolumn{1}{c}{\multirow{2}{*}{$\in [17, 123]$}} & \multicolumn{1}{l}{\multirow{2}{*}{mm}} &  & \multicolumn{1}{c}{\multirow{2}{*}{\begin{tabular}[l]{@{}l@{}}derived\\from \eqref{eq:d}\end{tabular}}} \\ 
			$d_2$ &  &  &  & \multicolumn{1}{c}{} \\
			$w$ & 3, 4, 5 & mm & 3 &  \\
			$s$ & 0.1, 0.3, 0.5 & mm & 3 & \\
			$N_T$ & 6, 8, 10 & turns & 3 & \\
			$N_L$ & 1, 2, 3, 4 & layers & 4 & \\
			$O$ & 0.5, 1.0, 1.5 & mm & 3 & \begin{tabular}[l]{@{}l@{}}not defined \\for $N_L = 1$\end{tabular}\\ [1.4ex]
			\hline\hline
		\end{tabular}
		\begin{tablenotes}
			\item[1] start:step:stop
		\end{tablenotes}
		\end{threeparttable}
	%	}
\end{table}

It should be noted that, using the geometric features of Table \ref{tab:dims}, the previous study \cite{estimation_pap25} introduced a new variable $O^\prime \equiv O^{N_L-1}$, i.e., the distance between layers $O$, raised to $(N_L-1)$, to account for the undefined state where $N_L=1$. Furthermore, following the current sheet approximation (CSA) principles established in historical methods \cite{monomial_original}, the geometric mean perimeters $\bar{D}_1 = (D_1+d_1)/2$ and $\bar{D}_2 = (D_2+d_2)/2$ were also introduced to track the path of the primary mutual flux linkages. Based on these, we construct a FEA dataset comprising 9 features, namely $D_1$, $D_2$, $\bar{D}_1$, $\bar{D}_2$, $w$, $s$, $N_T$, $N_L$, $O^\prime$, as well as the inductance $L$, as the target variable.

Generating a large-scale dataset via FEA is computationally expensive, often requiring months of numerical solver runtime. To optimize this process, we exploit the geometric symmetry of the problem. Because the spatial orientation of a rectangular winding does not affect its flux path or total self-inductance, a configuration with outer boundaries $\{D_1^*, D_2^*\}$ behaves identically to one with $\{D_2^*, D_1^*\}$. Enforcing the strict condition $D_1 \le D_2$ eliminates redundant combinations and cuts the required FEA simulation volume exactly in half.

For the generation of the FEA dataset, 3D parametric models within ANSYS Maxwell 3D are utilized. To ensure high numerical accuracy while optimizing computational efficiency, an auto-adaptive meshing scheme is applied across the global region, supplemented by an explicitly defined fine mesh localized within the immediate vicinity of the winding. This strategy yields a highly dense discretization ranging from approximately 700,000 to 1,000,000 tetrahedra per simulation. The iterative solver convergence process runs continuously until the global energy error drops below a strict 1\% threshold.

The complete open-source dataset developed for this work \cite{zenodo_db} is split into two parts: a core dataset containing 10,110 samples and an independent OOD testing dataset consisting of 1,124 samples. While the core dataset can be randomly partitioned into standard training and testing subsets to evaluate the interpolation accuracy of ML models, the OOD dataset is reserved exclusively to evaluate how well the models generalize to data from an entirely different distribution. The samples of the OOD dataset are categorized into seven classes:

\iffalse

\begin{itemize}
	\item Class 1: low-turn extension ($N_T = 4$) -- 120 samples
	\item Class 2: geometric footprint extension ($D_1, D_2 \ge 160$~mm) -- 360 samples
	\item Class 3: larger layer stack ($N_L = 6$) -- 170 samples
	\item Class 4: combined large footprint and layer configuration ($D_1, D_2 \ge 160$~mm and $N_L = 6$) -- 60 samples
	\item Class 5: ultra-wide track and spacing variant ($w = s = 1$~mm) -- 180 samples
	\item Class 6: extreme copper trace width ($w = 7$~mm) -- 84 samples
	\item Class 7: constricted inner aperture ($d_1 \le 17$~mm) -- 150 samples
\end{itemize}

\fi

\begin{itemize}
	\item low-turn extension ($N_T = 4$) -- 120 samples (Class 1)
	\item geometric footprint extension ($D_1, D_2 \ge 160$~mm) -- 360 samples (Class 2)
	\item larger layer stack ($N_L = 6$) -- 170 samples (Class 3)
	\item combined large footprint and layer configuration ($D_1, D_2 \ge 160$~mm and $N_L = 6$) -- 60 samples (Class 4)
	\item ultra-wide track and spacing variant ($w = s = 1$~mm) -- 180 samples (Class 5)
	\item extreme copper trace width ($w = 7$~mm) -- 84 samples (Class 6)
	\item constricted inner aperture ($d_1 \le 17$~mm) -- 150 samples (Class 7)
\end{itemize}

\section{Explainable Machine Learning Methodology} \label{sec3}

% % Spyros

To estimate the planar winding inductance $L$ from the $9$-dimensional geometric feature space detailed in Section \ref{sec2}, and to quantify the contribution of each feature to the resulting prediction, we implement an explainable framework structured into two distinct branches. The first branch applies classical ML algorithms to evaluate interpolation accuracy and utilizes post-hoc feature attribution to quantify feature importance. The second branch implements symbolic regression to derive closed-form analytical equations. To evaluate both interpolation fidelity and extrapolation robustness, all models are trained and tested under two partitioning schemes: a standard 80\%--20\% random split of the core dataset (hereafter \emph{core split}) and an out-of-distribution (OOD) extrapolation split, where models are trained on the full core dataset and evaluated exclusively on the OOD samples (hereafter \emph{OOD split}).

\subsection{Machine Learning and Post-Hoc Explainability} \label{sec3.1}

Across both data splits, the ML algorithms we deploy are Random Forest (RF) ensembles, Extreme Gradient Boosting (XGBoost), and Support Vector Regression (SVRs) with radial basis function kernels. In principle, RF constructs an ensemble of independent decision trees and averages their predictions to reduce variance \cite{RFs}, XGBoost sequentially trains weak base learners by optimizing a gradient-based objective function with regularization \cite{xgboost} and SVR maps inputs into a high-dimensional space to identify a linear boundary maximizing a specified margin tolerance \cite{SVR}. Prior to training, all input geometric features $\mathbf{x}$ and the target inductance $L$ are uniformly mapped to a logarithmic space via $\mathbf{x} \to \log_{10}(\mathbf{x} + \epsilon)$, where $\epsilon = 10^{-8}$ acts as a numerical stabilizer. Alternative transformation strategies standard in tabular ML, such as min-max scaling and standardization, were also evaluated; however, the uniform logarithmic transformation demonstrated superior numerical robustness across these models as well as all subsequent symbolic experiments.

The aforementioned models aim to minimize estimation error for $L$, therefore the use of post-hoc attribution methods that identify which geometric features are most meaningful to the predictions is required. To this end, we apply permutation importance, which quantifies global feature dependency by calculating the mean decrease in the coefficient of determination ($\Delta R^2$) when a given feature vector is randomly shuffled over $k$ iterations ($k=20$ in this study) \cite{permutation}, as well as SHAP, which computes local, additive feature attributions based on cooperative game theory \cite{shap}. To consolidate these attribution techniques, we compute a normalized consensus ranking ($CR$) for each feature. The individual importance scores from permutation importance and SHAP, denoted as $\bar{I}_{\text{perm}}$ and $\bar{I}_{\text{SHAP}}$ respectively, are first normalized and then averaged across all three models:

\begin{equation} \label{eq:cr}
	CR = \frac{1}{3} \sum_{m \in \mathcal{M}} \left( \frac{\bar{I}_{\text{perm}, m} + \bar{I}_{\text{SHAP}, m}}{2} \right)
\end{equation}

\noindent where $\mathcal{M} = \{\text{RF}, \text{XGBoost}, \text{SVR}\}$. This aggregated metric unifies the local attributions of SHAP with the global sensitivity of permutation importance into a single score.

\subsection{Symbolic Regression} \label{sec3.2}

Our previous work \cite{estimation_pap25} utilized MLR to fit a linear hyperplane to the log-transformed features of the original dataset, which enforces a rigid power-law monomial structure. While this approach builds upon prior work for the case of MLRPWs \cite{monomial_original}, enforcing a fixed symbolic form introduces structural bias that limits generalization performance across diverse application cases. To achieve problem-agnostic equation discovery without prior assumptions, this work primarily deploys the SR-KAN framework \cite{SRKAN}. SR-KAN addresses the bottlenecks of original Kolmogorov--Arnold Networks (KANs) for symbolic regression by replacing B-splines with computationally efficient radial basis functions parameterizing a Reflectional Switch Activation Function (RSWAF). Furthermore, to overcome the inability of standard additive KAN layers to capture product-based parameter couplings, the architecture explicitly incorporates separable multiplicative subunits \cite{KAN2}. Training is guided by an objective function that couples an $\ell_1$ magnitude penalty with row-wise and column-wise entropy regularizations to promote network sparsity. The operational pipeline executes a hierarchical search starting from single-layer, single-unit configurations up to deeper architectures, systematically pruning inactive edges before regressing the remaining univariate transformations against a symbolic function vocabulary to turn the network into a single analytic expression.

To evaluate the performance achieved by the expression derived by SR-KAN, we benchmark it against both the classical MLR formulation and PySR, a genetic programming framework widely recognized as the literature standard for symbolic regression. PySR explores a problem-agnostic algebraic search space composed of fundamental operators (e.g., $+$, $-$, $\times$, $\div$) via tournament selection and regularized mutations \cite{pysr}. Rather than optimizing a single isolated expression, both PySR and SR-KAN discover a multi-objective Pareto front of candidate formulas that balance mathematical complexity against predictive accuracy. Regarding accuracy, because the raw inductance values are inherently small, standard mean squared error metrics become numerically insensitive; therefore, after evaluating the candidate expressions in the log-transformed space, predictions are mapped back to the linear domain via a base-10 exponential transformation to compute the mean absolute relative error ($\mathcal{E}$):

\begin{equation} \label{eq:mre}
	\mathcal{E} = \frac{1}{N} \sum_{i=1}^{N} \left| \frac{L_{\text{pred}, i} - L_{\text{target}, i}}{L_{\text{target}, i}} \right|
\end{equation}

\noindent where $N$ denotes the number of samples, $L_{\text{target}}$ represents the true inductance from the FEA dataset, and $L_{\text{pred}}$ is the model prediction.

\section{Results \& Discussion} \label{sec4}

% Spyros mainly

This section presents the experimental results of the explainable framework proposed herein, deployed on the newly introduced MLRPW dataset. To ensure statistical significance against sources of inherent numerical randomness, such as the data partitioning of the core split or the weight initialization seeds of the neural layers within SR-KAN, all relevant experiments are executed across three independent random seeds. Performance metrics are reported as the mean value bounded by the standard error of the mean ($\pm \text{SEM}$). Throughout these benchmarks, all models are systematically compared against the baseline analytical expression derived in \cite{estimation_pap25}:

\begin{align} \label{eq:prev_mlr}
	L = &-5.7 - 0.59 D_1 - 0.38 D_2 + 1.18 \bar{D}_1 + 1.07 \bar{D}_2 \nonumber \\
	&- 0.18 w - 0.01 s + 1.79 N_T + 1.8 N_L - 0.006 O^\prime,
\end{align}

\noindent where the target inductance and all geometric features are log-transformed, but are written without the explicit functional notation of the logarithm for notational brevity. This convention is maintained for the remainder of the text, mapping the multiplicative power-law monomial directly into a linear expression.

\subsection{Machine Learning Results} \label{sec4.1}

Prior to executing the experiments, the hyperparameters for each regressor are tuned as follows: for the RF model, the ensemble size is configured to 300 independent decision trees to ensure variance reduction, with the maximum tree depth bounded at 20 to restrict individual estimator complexity. The XGBoost model is similarly tuned to 300 sequential base estimators utilizing the histogram-based tree method, which bin continuous features into discrete bins to significantly accelerate training throughput on tabular datasets. The SVR model is implemented using a radial basis function kernel, with the regularization parameter set to $C=100$ to penalize margin violations, a tube insensitivity threshold of $\epsilon=0.01$ to ignore minor training residuals and a data-dependent scaling parameter to modulate the decision boundary sensitivity. These hyperparameter selections are justified by the dimensionality of the feature space and total number of samples in the simulation dataset, based on common practices in tabular ML \cite{tab_ml}. The RF and SVR architectures are implemented via the \texttt{scikit-learn} Python library \cite{sklearn}, while XGBoost is deployed using the official \texttt{xgboost} library \cite{xgboost}.

The evaluation metrics achieved by these ML models across both data splits are presented in Table \ref{tab:ml}, which also includes the predictive performance of the reference analytical expression defined in Eq. \eqref{eq:prev_mlr}. Performance across both the core split and the OOD split is quantified via the mean coefficient of determination ($R^2$) alongside the mean absolute relative error ($\mathcal{E}$) defined in Eq. \eqref{eq:mre}. To reflect statistical significance, the relative error $\mathcal{E}$ is bounded explicitly by its corresponding standard error of the mean ($\pm\text{SEM}$) across the independent evaluation seeds for the core split. For the OOD split, the standard error reduces to zero or near-zero because the deterministic SVR formulation is seed independent, while the full-sample training routine over the entire core dataset stabilizes the tree-based ensembles against variance induced by subset partitioning (the core split is deterministic).

\begin{table}[t!]
	\renewcommand{\arraystretch}{1.3}
	\caption{Predictive Accuracy Comparison of Machine Learning Models Within the Core and OOD Splits}
	\centering
	\label{tab:ml}
	\resizebox{\columnwidth}{!}{
		\begin{tabular}{c c c c c}
			\hline\hline \\[-3mm]
			& \multicolumn{2}{c}{\textbf{Core Split}} & \multicolumn{2}{c}{\textbf{OOD Split}} \\ \cline{2-3} \cline{4-5} \\[-3mm]
			\multicolumn{1}{c}{\textbf{Model}} & $\mathbf{R^2}$ \textbf{(\%)} & $\boldsymbol{\mathcal{E}}$ \textbf{(\%)} & $\mathbf{R^2}$ \textbf{(\%)} & $\boldsymbol{\mathcal{E}}$ \textbf{(\%)} \\[0.5ex] \hline \\[-3mm]
			RF                  & $99.90$ & $1.87 \pm 0.02$ & $52.53$ & $36.65 \pm 0.00$ \\
			XGBoost   & $99.94$ & $1.43 \pm 0.02$ & $51.29$ & $37.45 \pm 0.00$ \\ [0.5ex]
			SVR        & $99.66$ & $3.08 \pm 0.05$ & $94.18$ & $12.95 \pm 0.00$ \\
			Reference MLR \cite{estimation_pap25} & $98.71$ & $5.88 \pm 0.17$ & $98.10$ & $9.92 \pm 0.00$ \\ [1.4ex]
			\hline\hline
		\end{tabular}
	}
\end{table}

The empirical results compiled in Table \ref{tab:ml} reveal distinct performance trade-offs between internal dataset interpolation and out-of-distribution extrapolation. Within the core split, all models achieve high predictive accuracy, yielding $R^2$ scores above $98$\%. The tree-based ensembles demonstrate the highest performance, with XGBoost being the leader ($\mathcal{E} = 1.43\%$), followed closely by RF. The non-linear SVR occupies a middle tier, whereas the reference monomial model yields the highest error rate within this group, approaching a mean absolute relative error of nearly $5.88\%$. This situation flips entirely for the OOD split. In this case, the reference monomial model maintains the lowest overall error profile ($\mathcal{E} = 9.92\%$), followed by the SVR model with an $R^2$ of $94.18$ and $\mathcal{E}= 12.95\%$. Conversely, the tree-based models experience severe performance degradation; their $R^2$ coefficients plummet to approximately $51\%$--$53\%$, while their mean absolute relative errors surge past $36\%$. This failure stems directly from the nature of recursive partitioning trees, which lack the capacity to extrapolate outside their learned feature bounds and are highly prone to overfitting.

\subsection{Attribution via SHAP and Permutation Importance} \label{sec4.2}

\begin{figure}[!b]\centering
	\includegraphics[width=0.8\columnwidth]{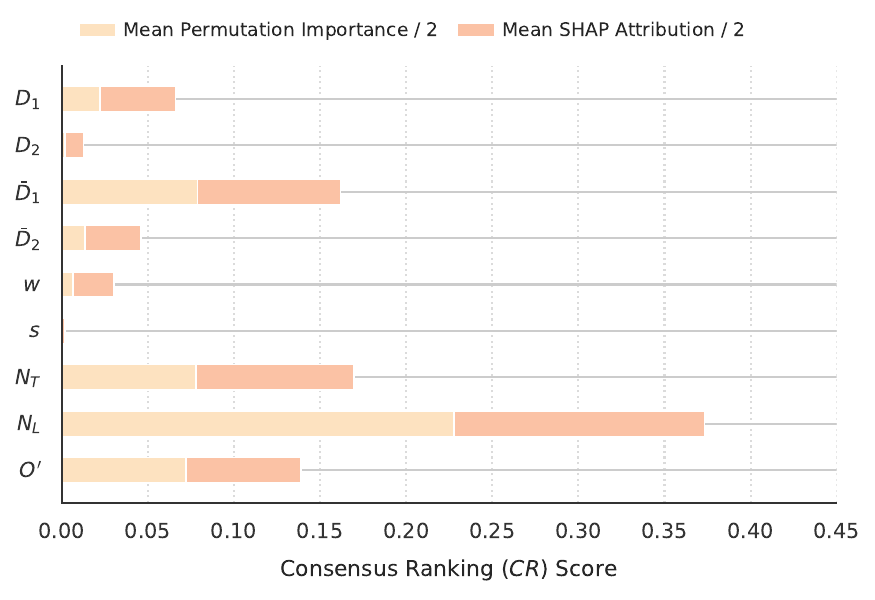}
	\caption{Consensus Ranking ($CR$) scores across the 9-dimensional MLRPW geometric space, showing the stacked contribution of mean permutation importance and mean SHAP attribution.}
	\label{fig:cr}
\end{figure}

To evaluate which features dominate the model decisions that lead to the performances presented in Table \ref{tab:ml}, post-hoc feature attribution is applied for all studied ML models. To this end, global feature variations are captured via permutation importance implemented using \texttt{scikit-learn}. Local, additive contributions are evaluated using the \texttt{shap} library \cite{shap}, employing the specialized \texttt{TreeExplainer} algorithm for the tree-based models \cite{shaptree} and a sample-averaged \texttt{KernelExplainer} approximation for SVR. To balance out the distinct metrics generated by these different techniques, the raw importance scores from each model are normalized so that the feature allocations within each individual method sum up to unity. The resulting relative feature metrics are summarized across models in Table \ref{tab:attr}, while the consolidated consensus ranking computed via Eq. \eqref{eq:cr} is illustrated in Fig. \ref{fig:cr}.

\begin{table}[t!]
	\renewcommand{\arraystretch}{1.3}
	\caption{Normalized Feature Importance Scores Across Machine Learning Models}
	\centering
	\label{tab:attr}
	\resizebox{\columnwidth}{!}{
		\begin{tabular}{c ccc ccc}
			\hline\hline \\[-3mm]
			& \multicolumn{3}{c}{\textbf{Permutation Importance}} & \multicolumn{3}{c}{\textbf{SHAP}} \\ \cline{2-4} \cline{5-7} \\[-3mm]
			\multicolumn{1}{c}{\textbf{Feature}} & \textbf{RF} & \textbf{XGBoost} & \textbf{SVR} & \textbf{RF} & \textbf{XGBoost} & \textbf{SVR} \\[0.5ex] \hline \\[-3mm]
			$D_1$         & $0.0150$ & $0.0087$ & $0.1030$ & $0.0769$ & $0.0482$ & $0.1350$ \\
			$D_2$         & $0.0004$ & $0.0028$ & $0.0071$ & $0.0044$ & $0.0258$ & $0.0333$ \\
			$\bar{D}_1$   & $0.0823$ & $0.0609$ & $0.3152$ & $0.1135$ & $0.1326$ & $0.2459$ \\
			$\bar{D}_2$   & $0.0482$ & $0.0234$ & $0.0109$ & $0.0881$ & $0.0679$ & $0.0407$ \\
			$w$           & $0.0164$ & $0.0210$ & $0.0027$ & $0.0483$ & $0.0733$ & $0.0220$ \\ [0.5ex]
			$s$           & $0.0002$ & $0.0004$ & $0.0001$ & $0.0031$ & $0.0066$ & $0.0014$ \\
			$N_T$         & $0.1980$ & $0.1156$ & $0.1561$ & $0.2008$ & $0.1827$ & $0.1702$ \\
			$N_L$         & $0.2675$ & $0.7651$ & $0.3246$ & $0.1908$ & $0.4399$ & $0.2465$ \\
			$O'$          & $0.3719$ & $0.0021$ & $0.0802$ & $0.2741$ & $0.0229$ & $0.1051$ \\ [1.4ex]
			\hline\hline
		\end{tabular}
	}
\end{table}

The relative normalized feature attributions compiled in Table \ref{tab:attr} demonstrate a strong consensus across all algorithms regarding parameter dominance. For all models and both attribution techniques, the macro-structural winding parameters -- specifically the total number of layers $N_L$ and the number of turns per layer $N_T$ -- emerge as the primary predictors. This effect is most pronounced in the XGBoost architecture, where $N_L$ accounts for $76.51\%$ of the global permutation drop and $43.99\%$ of the local SHAP attribution. Conversely, layout metrics such as the horizontal inter-turn trace spacing $s$ and the outer-side length $D_2$ consistently yield near-zero importance across all models.

The low individual attribution scores assigned to $D_2$ and the copper trace width $w$ do not indicate physical insignificance, but rather reflect the high multicollinearity inherent to Eq. \eqref{eq:d}. Because the uniform log-transformation preserves these linear dependencies, the models suffer from variance redundancy. Once an algorithm captures the global boundary via the primary outer-side length $D_1$ and the turn allocations ($N_T, N_L$), the variance explained by $D_2$ and $w$ becomes largely redundant, causing their isolated attribution scores to collapse. A similar mechanism is observed between the modified layer distance term $O^\prime$ and $N_L$. For the XGBoost model, the attribution metrics for $O^\prime$ appear practically negligible ($0.0021$ in permutation and $0.0229$ in SHAP); however, this is directly attributed to its structural relation to $N_L$, which acts as the dominant primary predictor and absorbs the shared variance during sequential gradient boosting.

The consolidated feature metrics are illustrated in Fig. \ref{fig:cr}, which explicitly displays the composition of the final consensus ranking as the superposition of both attribution methods. The visual breakdown confirms that $N_L$ is the heavily dominating structural driver across both local and global attribution methods, followed symmetrically by $N_T$ and $\bar{D}_1$. Conversely, features such as $s$ and $D_2$ show almost negligible individual segments, visually underscoring how collinearity suppresses their isolated ranking metrics. Ultimately, the fact that distinct machine learning architectures and completely different attribution techniques converge on this exact parameter hierarchy is highly significant; it uncovers the most influential features of the dataset, which are expected to be prioritized moving forward into the symbolic regression phase.

\subsection{Refining the MLR Approach} \label{sec4.3}

Before proceeding to fully unconstrained symbolic regression, we first revisit the MLR approach introduced in \cite{estimation_pap25}, which assumes a monomial expression for $L$. The previous study assumed that training on a subset of the current, full dataset was sufficient and that the full dataset would not yield significantly different model parameters. However, while the expression corresponding to Eq. \eqref{eq:prev_mlr} achieved a mean absolute relative error of 1.24\% in \cite{estimation_pap25}, its error increases to $5.88 \pm 0.17\%$ on the full dataset presented here. This nearly five-fold increase in error indicates that generating the complete dataset is highly meaningful; it also indicates that we should re-fit the MLR model to find an updated expression that performs better on the full dataset.

Following the logarithmic transformation and least-squares optimization methodology outlined in \cite{estimation_pap25}, the refined monomial expression for the MLRPW inductance is given by:

\begin{align}
	\label{eq:mlr}
	L = &- 5.5 -1.8D_1 + 0.11D_2 + 2.2\bar{D}_1 + 0.65\bar{D}_2 \nonumber \\
	&- 0.053w + 0.01s + 2.0N_T + 1.8N_L - 0.005O^\prime.
\end{align}

\noindent Evaluating this updated expression yields a mean absolute relative error of $\mathcal{E} = 4.12 \pm 0.03\%$ alongside a determination coefficient of $R^2 = 99.51\%$ for the core split. For the OOD split, the refined model achieves an error of $\mathcal{E} = 7.91\%$ and an $R^2$ of $98.98\%$. While this refinement does indeed lead to better performance, it does not alter the qualitative conclusions established in the previous benchmarks: the standard ML models of Section \ref{sec4.1} still exhibit superior interpolation within the core split, whereas the MLR model shows superior robustness when generalizing outside the domain bounds for the OOD split.

\begin{figure}[b!]\centering
	\includegraphics[width=\columnwidth]{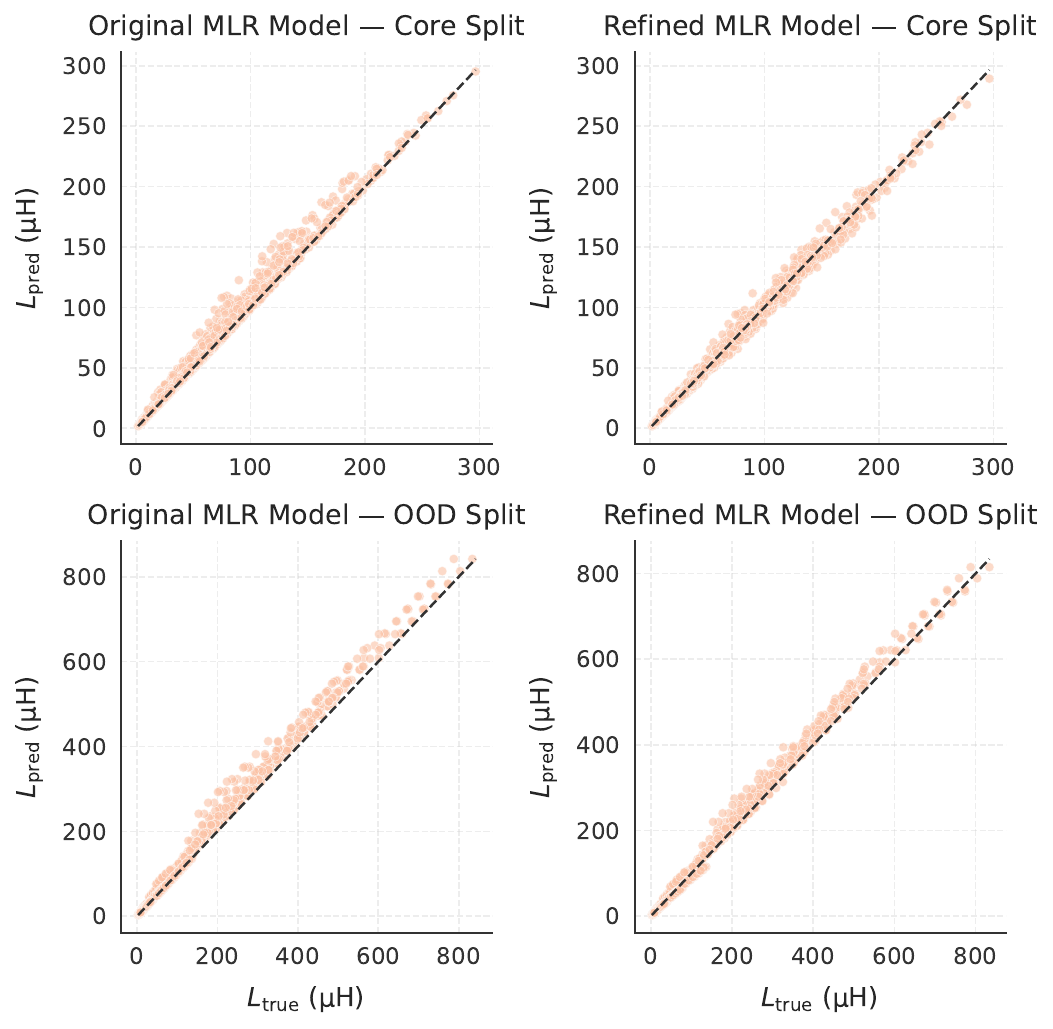}
	\caption{Comparison plots mapping the predicted self-inductance $L_{\text{pred}}$ against the true self-inductance $L_{\text{true}}$ for the original monomial (left column) and the refined MLR model (right column) evaluated across both the core split (top row) and the OOD split (bottom row). The dashed diagonal represents the ideal zero-error line ($L_{\text{pred}}=L_{\text{true}}$).}
	\label{fig:mlr_res}
\end{figure}

To illustrate the performance improvement, the plots comparing the dataset ground-truth inductance $L_{\text{true}}$ against the model predictions $L_{\text{pred}}$ are presented in Fig. \ref{fig:mlr_res}. Across both the core and OOD splits, the data points for the refined model of Eq. \eqref{eq:mlr} display a noticeably tighter concentration along the ideal $L_{\text{pred}}=L_{\text{true}}$ identity line. This reduction in dispersion is especially evident at higher inductance values (above 150~$\mu$H in the core split and above 400~$\mu$H in the OOD split) where the original expression consistently exhibits a systematic upward bias, overestimating the target inductance.

The sign and magnitude of the coefficients in Eqs. \eqref{eq:prev_mlr} and \eqref{eq:mlr} provide valuable physical insights, as they correspond directly to the power-law exponents of the underlying monomial. In the following analysis, the explicit $\log$ notation is re-introduced to avoid confusion between the physical geometric features and their log-transformed counterparts.

In both models, the coefficients for $\log N_T$ and $\log N_L$ have the highest relative magnitudes (excluding the geometric variables), confirming the primary importance ranking shown in the $CR$ analysis of Fig. \ref{fig:cr}. Note that the near-zero exponent of the modified layer distance $\log O^\prime$ ($\approx -0.005$) in both expressions is not an indication of physical insignificance, but rather a consequence of its high collinearity with $N_L$, which suppresses its individual weight.

However, an apparent discrepancy arises when observing the coefficient of $\log D_2$, which flips from $-0.38$ in Eq. \eqref{eq:prev_mlr} to $+0.11$ in Eq. \eqref{eq:mlr}. While a sign flip typically suggests a contradiction in the predicted physics of the system, this behavior is a mathematical artifact of the strong collinear coupling between $D_2$ and $\bar{D}_2 \equiv (D_2 + d_2)/2$. Following Eq. \eqref{eq:d}, $d_2$ can be expressed as $d_2 = D_2 - 2\Delta$, where $\Delta = N_T\left(w+s\right)- s$, therefore combining these two expressions yields:

\begin{equation}
	\label{eq:dbar}
	\bar{D}_2 = D_2 \left(1 - \frac{\Delta}{D_2}\right).
\end{equation}

\noindent As a first-order approximation, one may assume that $\Delta/D_2$ is sufficiently small, so that performing a Taylor expansion and keeping the linear terms, i.e., $\log \bar{D}_2  \approx \log D_2 - \Delta/D_2$, is justified. Then, the net effective coefficients for $\log D_2$ in Eqs. \eqref{eq:prev_mlr} and \eqref{eq:mlr} become $-0.38 + 1.07 = 0.69 > 0$ and $0.11 + 0.65 = 0.76 > 0$, respectively, demonstrating that the underlying physics predicted by both expressions is consistent.

\subsection{Symbolic Regression Results}  \label{sec4.4}

With the refined MLR model established as our new baseline, we proceed to fully unconstrained symbolic regression using SR-KAN alongside PySR, the current domain standard. For the PySR execution, we select 300 iterations across 50 populations, deploying an extended mathematical vocabulary that includes basic algebraic operators and the power operation. Each run generates a collection of candidate models from which the framework automatically elects the Pareto-optimal expression, balancing low functional complexity against high predictive accuracy.

\begin{table}[b!]
	\renewcommand{\arraystretch}{1.3}
	\caption{Predictive Accuracy Comparison of Symbolic Regression Frameworks Within the Core and OOD Splits}
	\centering
	\label{tab:sr}
	\resizebox{\columnwidth}{!}{
		\begin{tabular}{c c c c c}
			\hline\hline \\[-3mm]
			& \multicolumn{2}{c}{\textbf{Core Split}} & \multicolumn{2}{c}{\textbf{OOD Split}} \\ \cline{2-3} \cline{4-5} \\[-3mm]
			\multicolumn{1}{c}{\textbf{Model}} & $\mathbf{R^2}$ \textbf{(\%)} & $\boldsymbol{\mathcal{E}}$ \textbf{(\%)} & $\mathbf{R^2}$ \textbf{(\%)} & $\boldsymbol{\mathcal{E}}$ \textbf{(\%)} \\[0.5ex] \hline \\[-3mm]
			Refined MLR                         & $99.51$ & $4.12 \pm 0.03$ & $98.98$ & $7.91 \pm 0.00$ \\ [0.5ex]
			PySR                                & $99.29$ & $4.74 \pm 0.17$ & $96.14$ & $17.32 \pm 3.17$ \\
			SR-KAN                              & $99.49$ & $3.77 \pm 0.04$ & $98.42$ & $8.22 \pm 0.55$ \\ [1.4ex]
			\hline\hline
		\end{tabular}
	}
\end{table}

For the SR-KAN implementation, we enforce an absolute error threshold of 0.08 in the log-transformed space between prediction and target. The base regularization coefficient is set to $\lambda_0 = 2 \times 10^{-4}$, with relative regularization components configured as $\lambda_1 = 1$ for magnitude regularization, $\lambda_2 = 2$ for entropy regularization and $\lambda_3 = 1$ for $l_1$-regularization \cite{SRKAN}. The results for each framework are summarized in Table \ref{tab:sr}. Note that, because the OOD split is deterministic, the refined MLR baseline yields no standard error there; however, the stochastic nature of PySR and SR-KAN necessitates seed-averaging across both splits.

As demonstrated in Table \ref{tab:sr}, all three frameworks achieve high predictive accuracy within the core split, consistently yielding $R^2$ values above $99\%$. Within this domain, SR-KAN exhibits a slight performance edge with a mean absolute relative error of $3.77 \pm 0.04\%$, with variance as low as that of the refined MLR model. Conversely, PySR yields slightly lower accuracy and higher variance compared to both alternative frameworks. The performance divergence becomes far more pronounced when examining the results for the OOD split. Within this regime, the refined MLR baseline maintains the highest performance, followed very closely by SR-KAN. In contrast, PySR exhibits significantly degraded generalization capability, producing a mean error of $17.32 \pm 3.17\%$, which is more than double that of the other two models.

These results demonstrate that SR-KAN not only outperforms PySR in both splits, but also matches or exceeds the performance of the refined MLR baseline. This represents a major milestone: while the MLR framework relies on a rigid, pre-defined monomial assumption rooted in previous literature to constrain its optimization, SR-KAN achieves equivalent generalization without any prior assumptions regarding the underlying functional form. Consequently, this robustness highlights KAN-based symbolic regression as an exceptionally powerful tool for industrial electronics applications where the exact parametric scaling laws are  unknown, or impractical to determine.

To examine the specific geometric features selected by each framework and verify if they align with the consensus feature rankings established in Section \ref{sec4.2}, we isolate the expressions from the best-performing individual runs on the OOD split. The optimal PySR model achieves a determination coefficient of $R^2 = 99.31\%$ with an error of $\mathcal{E} = 4.46\%$ for the core split, alongside OOD metrics of $R^2 = 97.02\%$ and $\mathcal{E} = 13.84\%$, yielding the explicit functional form presented in Eq. \eqref{eq:pysr}.

\begin{align}
	\label{eq:pysr}
	L = - 2.69 + N_L + N_T^{1.42} + \frac{D_2 + N_L + \bar{D}_1}{1.15} + \frac{3.72}{w}
\end{align}

\noindent Conversely, the best-performing SR-KAN expression achieves a core-split performance of $R^2 = 99.46\%$ and $\mathcal{E} = 3.84\%$, while maintaining remarkable generalization across the OOD split with $R^2 = 98.73\%$ and $\mathcal{E} = 7.36\%$ (even lower than the refined MLR baseline). This optimal analytical expression is provided in Eq. \eqref{eq:srkan}.

\begin{align}
	\label{eq:srkan}
	L = &- 2.58 - 0.46\bar{D}_1^2 + 1.2\bar{D}_1 + 0.74\bar{D}_2 + 2.05N_T \nonumber \\
	&- 0.77\sqrt{10.3 D_1 + 14.66}  \\
	& + 2.1\sin(0.92N_L) - 0.02\sin(6.41N_L) \nonumber
\end{align}

An inspection of the discovered equations reveals that both symbolic frameworks select $N_L$, $N_T$, and $\bar{D}_1$. These variables exactly mirror three of the top four parameters identified by the consensus ranking in Section \ref{sec4.2}. While the fourth high-ranking parameter, $O^\prime$, is omitted by both algorithms, its exclusion follows the exact collinearity mechanism observed in the MLR baseline; its high consensus ranking during post-hoc feature attribution stems from its coupling with $N_L$. Additionally, both models selectively retain a sparse subset of minor geometric metrics: $D_2$ and $w$ for PySR and $\bar{D}_2$ and $D_1$ for SR-KAN. This demonstrates another advantage of unconstrained symbolic regression, as lower-importance features (such as $s$) are systematically pruned, leading to simpler expressions.

\begin{figure}[t!]\centering
	\includegraphics[width=\columnwidth]{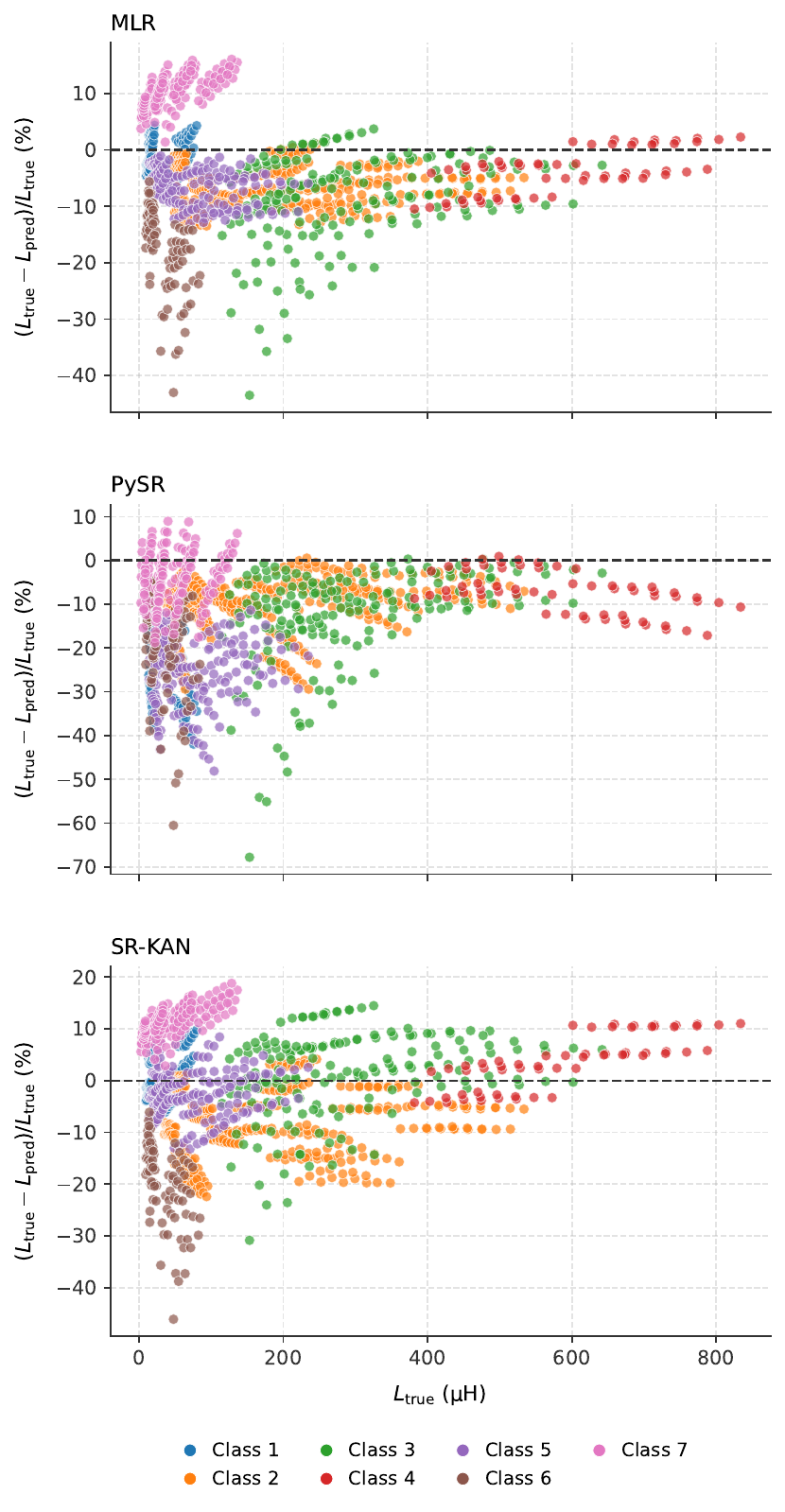}
	\caption{Relative residuals $(L_{\text{true}} - L_{\text{pred}}) / L_{\text{true}}$ plotted against the target self-inductance $L_{\text{true}}$ on the OOD split for the refined MLR, PySR and SR-KAN models. The individual error points are color-coded according to the seven distinct OOD classes.}
	\label{fig:ood_res}
\end{figure}

To conclude this analysis, we provide in Fig. \ref{fig:ood_res} the relative residuals for the refined MLR, PySR and SR-KAN expressions evaluated exclusively on the OOD split. To isolate exactly where each expression struggles or excels, the individual residuals are color-coded according to the seven distinct OOD classes defined in the end of Section \ref{sec2}. A primary observation is the distinct performance degradation of the PySR framework, which exhibits a significantly broader vertical dispersion, with relative errors reaching as high as $70\%$ in magnitude for specific edge cases (Class 3). Conversely, the refined MLR and SR-KAN expressions display a far more constricted residual range, bounding the majority of their tracking errors within a much tighter envelope. Furthermore, MLR and SR-KAN demonstrate remarkably congruent error characteristics across the individual geometric classes; both frameworks share very similar distribution topologies for most classes, which are nearly identical in the cases of Class 6 and Class 7. In contrast, the PySR residual profile behaves independently. This cross-framework structural alignment strongly suggests that SR-KAN independently discovered an analytical mapping that mirrors the foundational power-law scaling principles embedded within the MLR formulation, accounting for its stable generalization across the OOD split.

\section{Experimental Validation} \label{sec5}

% % Theofilos

To assess the numerical accuracy and physical validity of the parametric FEA models, experimental validation was performed on hardware prototypes. Over 50 distinct MLRPW specimens were fabricated using standard commercial PCB prototyping techniques. The manufactured component portfolio covers a broad geometric design landscape, featuring trace widths up to 7~mm, layer counts extending from single-layer up to an 8-layer stack and a wide experimental self-inductance range spanning from 4.11~$\mu$H to 559.27~$\mu$H. The physical parameter ranges and boundary metrics of these experimental test specimens are summarized in Table \ref{tab:experimental_summary}.

\begin{table}[!b]
	\renewcommand{\arraystretch}{1.2}
	\caption{Experimental Prototype Ranges and FEA Validation Summary (56 Samples)}
	\label{tab:experimental_summary}
	\centering
	\begin{tabular}{l c c c}
		\hline\hline \\[-3mm]
		\textbf{Physical Parameter} & \textbf{Minimum} & \textbf{Maximum} & \textbf{Unit} \\ 
		\hline \\[-3mm]
		Outer dimension $D_1$     & 70.0  & 210.0  & mm \\
		Outer dimension $D_2$     & 100.0 & 294.0  & mm \\
		Trace width $w$           & 3.0   & 7.0    & mm \\
		Trace spacing $s$         & 0.1   & 2.0    & mm \\
		Turns per layer $N_T$     & 4     & 12     & turns \\
		Layer count $N_L$         & 1     & 8      & layers \\
		Measured $L_{\text{meas}}$ & 4.11  & 559.27 & $\mu$H \\ 
		\hline \hline
	\end{tabular}
	
	\vspace{3mm}
	
	\begin{tabular}{l c c}
		\hline\hline \\[-3mm]
		\textbf{Error Metric ($L_{\text{sim}}$ vs. $L_{\text{meas}}$)} & \textbf{All Samples} & \textbf{Excl. Sample \#36} \\ 
		\hline \\[-3mm]
		Mean Relative Error (\%)       & $-0.757$ & $-0.228$ \\
		Mean Absolute Error (\%)       & $2.882$  & $2.392$  \\
		Standard Deviation $\sigma$ (\%) & $5.014$  & $3.104$  \\
		\hline \hline
	\end{tabular}
\end{table}

Laboratory measurements were conducted utilizing an HP 4284A high-precision LCR meter, which offers a manufacturer-specified accuracy ranging from $\pm0.1\%$ for high-impedance windings to $\pm1\%$ for low-impedance configurations. To suppress unwanted residual impedance, stray capacitance and induction errors introduced by the test leads, the OPEN and SHORT compensation calibration routines were executed immediately prior to data collection at a test frequency of $100$~kHz, matching the harmonic excitation of the FEA simulations.

% the original monomial (left column) and the refined MLR model (right column) evaluated across both the core split (top row) and the OOD split (bottom row). The dashed diagonal represents the ideal zero-error line ($L_{\text{pred}}=L_{\text{true}}$).

The tracking correlation between the numerical simulations and physical laboratory findings demonstrates exceptional consistency across the entire dataset. Evaluating all 55 prototype configurations yields a mean relative error of $-0.757\%$. A single prominent tracking anomaly is observed at Sample \#36, where the numerical solver underestimates the physical hardware response by $-29.86\%$. This localized deviation is classified as a physical manufacturing defect, likely originating from internal copper layer misalignment or localized prepreg thickness variations during the board pressing stage. Excluding this single anomalous specimen from the validation pool compresses the global mean relative error to a negligible $-0.228\%$, while achieving an overall mean absolute relative error of just $2.39\%$ and a tight standard deviation of $3.10\%$. These tight error bounds confirm that the synthetic FEA dataset reflects real-world physical behavior, establishing a mathematically sound and highly dependable framework for training and testing the explainable ML models and SR frameworks.

% To evaluate the practical utility of the derived closed-form expressions, Eqs. \eqref{eq:mlr} and \eqref{eq:srkan} are deployed directly as inductance estimators for the physical windings. Evaluating the 54 physical specimens (excluding \#36), the refined MLR monomial expression of Eq. \eqref{eq:mlr} achieves $\mathcal{E}= 4.58$\% with a standard deviation of 6.11\%. Crucially, the unconstrained expression of Eq. \eqref{eq:srkan} discovered via the SR-KAN framework achieves a highly competitive error of $\mathcal{E}= 6.26$\% and a standard deviation of 8.05\%. Both equations maintain tight prediction bounds even on extreme structural variants -- such as the 6-layer and 8-layer vertical stacks ($N_L \in \{6, 8\}$) -- confirming that KAN-guided equation discovery translates successfully from synthetic simulation environments to real-world power components.

\begin{figure}[tbh!]\centering
	\includegraphics[width=\columnwidth]{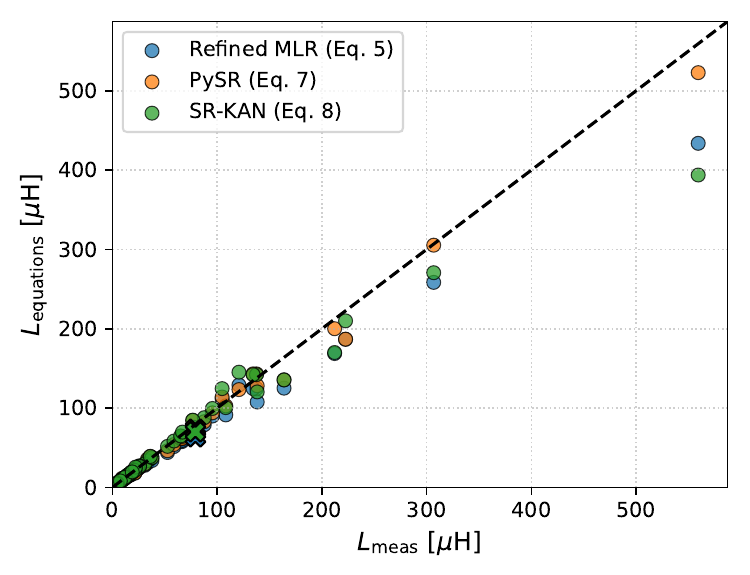}
	\caption{Comparison plots mapping the measured self-inductance $L_{\text{meas}}$ against the estimated self-inductance $L_{\text{equations}}$ for equations \eqref{eq:mlr}, \eqref{eq:pysr}, and \eqref{eq:srkan}.}
	\label{fig:exp_plot}
\end{figure}

To evaluate the practical utility of the derived closed-form expressions, Eqs. \eqref{eq:mlr}, \eqref{eq:pysr}, and \eqref{eq:srkan} are compared against the laboratory measurements $L_{\text{meas}}$. As illustrated in Fig.~\ref{fig:exp_plot}, all expressions demonstrate strong alignment across the entire experimental range. Evaluating the 55 valid physical specimens (excluding the Sample~\#36 defect), the refined MLR monomial \eqref{eq:mlr} achieves a mean absolute relative error of $\mathcal{E} = 7.11\%$ and a standard deviation of $\sigma = 7.38\%$. The PySR model \eqref{eq:pysr} achieves an accuracy of $\mathcal{E} = 6.10\%$ ($\sigma = 7.98\%$). The unconstrained expression discovered via the SR-KAN framework \eqref{eq:srkan} yields an error of $\mathcal{E} = 5.98\%$ ($\sigma = 8.33\%$). All three analytical expressions maintain reliable prediction bounds across demanding structural edge cases, such as $N_L \in \{6, 8\}$, confirming that symbolic regression translates successfully from synthetic FEA simulation environments to real-world power electronic hardware.

\section{Conclusion} \label{sec6}

% % Theofilos + Spyros

This paper introduced a holistic, explainable ML framework for estimating the inductance of MLRPWs across broad geometric domains. By unifying SHAP and permutation importance, as post-hoc feature attribution techniques, with the SR-KAN framework for symbolic regression, the proposed framework achieves high predictive accuracy while yielding transparent, closed-form mathematical equations. Unlike classical analytical models or prior MLR approaches that enforce a rigid power-law monomial structure, SR-KAN enables problem-agnostic equation discovery without requiring predefined functional assumptions about the underlying physical system. By removing these pre-enforced mathematical constraints, the framework yields a structurally unbiased equation discovery process that naturally captures complex non-linear parametric couplings.
%, other than forcing $\log_{10}(.)$ to the data and applying $O^\prime \equiv O^{N_L-1}$.

Benchmarking across both the core and OOD splits revealed critical trade-offs regarding model architecture and extrapolation capability. Standard tree-based algorithms (XGBoost and RF) achieved near-perfect interpolation within the core split ($\mathcal{E} < 1.9\%$), but suffered severe performance degradation when evaluated on OOD edge cases ($\mathcal{E} > 36\%$). This failure underscores that standard black-box regressors cannot be safely extrapolated for hardware optimization outside their explicit training boundaries. Conversely, the closed-form expression discovered by SR-KAN maintained stable performance, yielding an average OOD error of 8.22\% (with an optimal single run reaching 7.36\%), matching and occasionally exceeding the 7.91\% error of the refined MLR baseline. %This represents a clear progression from earlier linear regression methods, proving that KAN-guided symbolic search can extract foundational scaling laws directly from data.

Post-hoc feature attribution established a consistent parameter ranking across distinct ML architectures, identifying layer count, turns per layer, and geometric mean perimeters as the primary structural drivers of self-inductance. Furthermore, the symbolic regression models automatically pruned highly collinear or low-impact variables, such as inter-turn trace spacing, producing compact algebraic expressions that mirror CSA principles. While the inter-layer distance term $O^\prime$ was pruned by SR-KAN (and the PySR baseline) for the specific dataset due to its high correlation to $N_L$, its impact is anticipated to be far more dominant in WPT applications, where significantly larger $O^\prime$ geometry strongly governs magnetic field coupling and spatial field distribution.

The FEA simulation environment was experimentally validated against $55$ custom-fabricated PCB prototypes, spanning a wide range of inductances and layer counts. Excluding a single physical manufacturing defect, the FEA solver aligned closely with hardware measurements, exhibiting a mean relative error of $-0.23\%$ and a mean absolute relative error of $2.39$\% with a standard deviation of $3.10$\%. When evaluated directly to physical windings, the refined MLR monomial and the SR-KAN-discovered equation achieved mean absolute relative errors of $7.11$\% and $5.98$\%, respectively. Both formulas maintained tight tracking bounds across demanding physical edge cases, confirming that the equations translate effectively to real-world hardware.

Beyond planar magnetics, the proposed methodology offers a fully domain-agnostic explainable pipeline that can be readily generalized to any data-driven modeling or optimization dataset in power electronics, without being restricted to inductance estimation or planar components. To support further research in data-driven magnetics, the complete dataset of over 10,000 FEA simulations, alongside the physical prototype measurement database, has been made publicly available \cite{zenodo_db}, with detailed guidelines and routines for loading the data hosted in a public GitHub repository \cite{github_db}.

\iffalse

// Include a comparative table showing the results from your JESTIE (Regression) paper vs. the new KAN results.

// "No A Priori Assumptions" SR-KAN achieved good performance to OOD, matching MRL baseline (exceed for some parts?). 

// Standard techniques present good performance for ID, but terrible for OOD.

// FEA solver alignment with 56 physical prototypes (very small errors). 10,000+ FEA simulation dataset and 56 physical prototype measurements are released as open-source data.

this is an equation \eqref{eq:lamda}

\begin{align} \label{eq:lamda}
	\nonumber\mathbf \int_{0}^{{r}_2} & F(r,\varphi) dr \ d\varphi = [\sigma{r}_2 / (2{\mu}_0)]
	\\
	& \int_{0}^{\infty} exp(-\lambda|{z}_j - {z}_i|){\lambda}^{-1} {J}_1 (\lambda {r}_2) {J}_0 (\lambda {r}_i) d \lambda .
\end{align}

\begin{figure}[!t]\centering
	\includegraphics[width=8.5cm]{figures/FIG1.eps}
	\caption{Magnetization as a function of applied field. Note that ``Fig.'' is abbreviated. There is a period after the figure number, followed by two spaces. It is good practice to explain the significance of the figure in the caption.}
	\label{fig:fig_1}
\end{figure}

\section*{Appendix}

Appendixes, if needed, appear before the acknowledgment.

\fi 

% References

\bibliographystyle{Bibliography/IEEEtranTIE}
\bibliography{Bibliography/IEEEabrv,Bibliography/BIB_MLR_KAN}\ %IEEEabrv instead of IEEEfull

\end{document}